%% file: main.tex
\documentclass{article}

    \PassOptionsToPackage{numbers, compress}{natbib}

\usepackage[final]{neurips_2024_x}

\usepackage[utf8]{inputenc} 
\usepackage[T1]{fontenc}    
\usepackage{hyperref}       
\usepackage{url}            
\usepackage{booktabs}       
\usepackage{amsfonts}       
\usepackage{nicefrac}       
\usepackage{microtype}      
\usepackage{xcolor}         

\usepackage{hyperref}
\usepackage{url}

\usepackage{graphicx}
\usepackage{amsmath}
\usepackage{amssymb}

\usepackage{gensymb}

\usepackage{amssymb}
\usepackage{xcolor}

\usepackage{subcaption}

\usepackage{kotex}
\usepackage{color}
\usepackage{array}

\usepackage{bm}
\usepackage{nth}

\newcolumntype{C}[1]{>{\centering\let\newline\\\arraybackslash\hspace{0pt}}m{#1}}
\newcolumntype{L}[1]{>{\let\newline\\\arraybackslash\hspace{0pt}}m{#1}}

\usepackage{rotating} 

\usepackage{amsmath}

\usepackage{booktabs}
\usepackage{listings}
\usepackage{algorithm}
\usepackage{algorithmic}
\usepackage{mathtools, nccmath}

\usepackage[capitalize]{cleveref}
\crefname{section}{Sec.}{Secs.}
\Crefname{section}{Section}{Sections}
\Crefname{table}{Table}{Tables}
\crefname{table}{Tab.}{Tabs.}

\title{GenMix: Combining Generative and Mixture Data Augmentation for Medical Image Classification}

\author{%
  Hansang~Lee
  \quad
  Haeil~Lee \\
  School of Electrical Engineering\\
  Korea Advanced Institute of Science and Technology\\
  Daehark 291, Yuseonggu, Daejeon 34141, Republic of Korea\\
  \texttt{\{hansanglee,haeil.lee\}@kaist.ac.kr}\\
  \And
  Helen~Hong~\thanks{Corresponding author} \\
  Department of Software Convergence \\
  Seoul Women’s University \\
  Hwarangro 621, Nowongu, Seoul 01797, Republic of Korea \\
  \texttt{hlhong@swu.ac.kr} \\
}

\begin{document}

\maketitle

\input{sections/0_abstract}    
\input{sections/1_introduction}
\input{sections/3_methods}
\input{sections/4_experiments}
\input{sections/6_conclusion}

\subsubsection*{Acknowledgments}
This research was supported by Basic Science Research Program through the National Research Foundation of Korea (NRF) funded by the Ministry of Education (2022R1I1A1A01071970), and the National Research Foundation of Korea Grant funded by the Korea government (No. RS-2023-00207947).

\bibliography{main}
\bibliographystyle{ieeetr}


\end{document}

%% file: sections/0_abstract.tex
\begin{abstract}
In this paper, we propose a novel data augmentation technique called GenMix, which combines generative and mixture approaches to leverage the strengths of both methods. 
While generative models excel at creating new data patterns, they face challenges such as mode collapse in GANs and difficulties in training diffusion models, especially with limited medical imaging data. 
On the other hand, mixture models enhance class boundary regions but tend to favor the major class in scenarios with class imbalance. 
To address these limitations, GenMix integrates both approaches to complement each other.
GenMix operates in two stages: (1) training a generative model to produce synthetic images, and (2) performing mixup between synthetic and real data. 
This process improves the quality and diversity of synthetic data while simultaneously benefiting from the new pattern learning of generative models and the boundary enhancement of mixture models. 
We validate the effectiveness of our method on the task of classifying focal liver lesions (FLLs) in CT images.
Our results demonstrate that GenMix enhances the performance of various generative models, including DCGAN, StyleGAN, Textual Inversion, and Diffusion Models. 
Notably, the proposed method with Textual Inversion outperforms other methods without fine-tuning diffusion model on the FLL dataset. 
\end{abstract}

%% file: sections/1_introduction.tex
\section{Introduction}

Data augmentation (DA) plays a crucial role in medical image analysis, significantly enhancing the performance of machine learning models by increasing the diversity and volume of training data~\citep{Shorten2019}. 
This is particularly important in medical imaging, where acquiring large and varied datasets is often challenging due to privacy concerns, high costs, and the rarity of certain conditions. 
Effective DA techniques can help mitigate issues such as over-fitting and class imbalance, thereby improving model generalization and robustness.
However, standard augmentation methods such as rotation, scaling, and flipping may not be sufficient to capture the complex variations and subtle differences present in medical images, necessitating the development of more sophisticated approaches.

In recent years, generative models like Generative Adversarial Networks (GANs) and Diffusion Models have gained popularity in medical image analysis for their ability to create realistic synthetic data. 
These models can generate new patterns that resemble real medical images, thus expanding the training dataset. 
For instance, Salehinejad et al. used Deep Convolutional GAN (DCGAN)~\citep{Radford2015DCGAN} and AlexNet~\citep{Krizhevsky2012AlexNet} to improve classification performance in the task of identifying five diseases of chest X-ray images~\citep{salehinejad2018generalization}. 
Frid-Adar et al. employed DCGAN and Auxiliary Conditional GAN (ACGAN)~\citep{Odena2016ACGAN} to classify liver lesions in abdominal CT images using LeNet-like network~\citep{frid2018gan}. 
Zhao et al. proposed Forward and Background GAN (F\&BGAN) to classify lung nodules in chest CT images using modified VGG-16 network~\citep{zhao2018synthetic}. 
Lee et al. used DCGAN and pix2pix~\citep{Isola2016pix2pix} to classify focal liver lesions in abdominal CT images using AlexNet~\citep{Lee2021}.
Generative models offer the advantage of generating high-quality synthetic data of novel patterns, which can be more diverse and representative of real-world variability compared to transform-based data augmentation techniques. 
However, generative models often face challenges such as mode collapse in GANs and the complexity of training diffusion models, especially when the amount of available training data is limited. 

Alternatively, mixture models, which combine data from different classes, have been used to enhance class boundary regions, making models more robust to variations. 
For instance, Nishio et al. used MixUp~\citep{Zhang2017MixUp} to classify COVID-19, pneumonia, and normal chest X-ray images using the VGG-16 network~\citep{nishio2020automatic}. 
Rahan et al. applied MixUp to classify five diseases in chest X-ray images using the ResNet-18 network~\citep{rajan2021selftraining}.
Özdemir et al. employed MixUp to distinguish COVID-19 from normal chest CT images using the ResNet-50 and ResNet-101 networks~\citep{ozdemir2022attention}. 
Mixture models have been shown to improve the generalization and robustness of deep learning models by diversifying the patterns of training data despite their simple calculations.
However, mixture models can become biased towards major classes in scenarios of class imbalance, leading to suboptimal performance.

To address the limitations of existing DA techniques, we propose a novel approach called GenMix. 
GenMix synergistically combines the strengths of generative and mixture models, leveraging the benefits of both methodologies while mitigating their individual shortcomings. 
Our approach consists of two main stages: (1) training a generative model to produce synthetic images, and (2) performing mixup between synthetic and real data. 
By integrating these steps, GenMix improves the quality and diversity of synthetic data, while simultaneously enhancing the learning of new data patterns from generative models and the boundary strengthening properties of mixture models.
The experimental results demonstrate that the proposed GenMix consistently improves the effectiveness of data augmentation across different types of generative models. 
Moreover, a type of diffusion model known as Textual Inversion~\citep{Gal2023TextualInversion} achieved the best classification performance on the FLL dataset without any fine-tuning, solely by applying GenMix.

Our contributions are as follows.
\begin{itemize}
    \item We introduce GenMix, a novel data augmentation technique that effectively combines generative and mixture approaches to improve the performance of medical image classification models.
    \item We demonstrate the effectiveness of GenMix in the task of classifying focal liver lesions (FLL) in CT images, showcasing its ability to handle challenges such as mode collapse and class imbalance.
    \item We validate our method across various generative models, including DCGAN, StyleGAN, Textual Inversion, and Diffusion Models, and show consistent improvements in generative model efficiency.
    \item We highlight the notable performance of Textual Inversion, which surpasses other methods without requiring fine-tuning on the FLL dataset, underscoring the efficacy of our proposed approach.
\end{itemize}

%% file: sections/3_methods.tex
\section{Methods}
\label{sec:methods}

\begin{figure*}[t!]
\centering
\includegraphics[width=\textwidth]{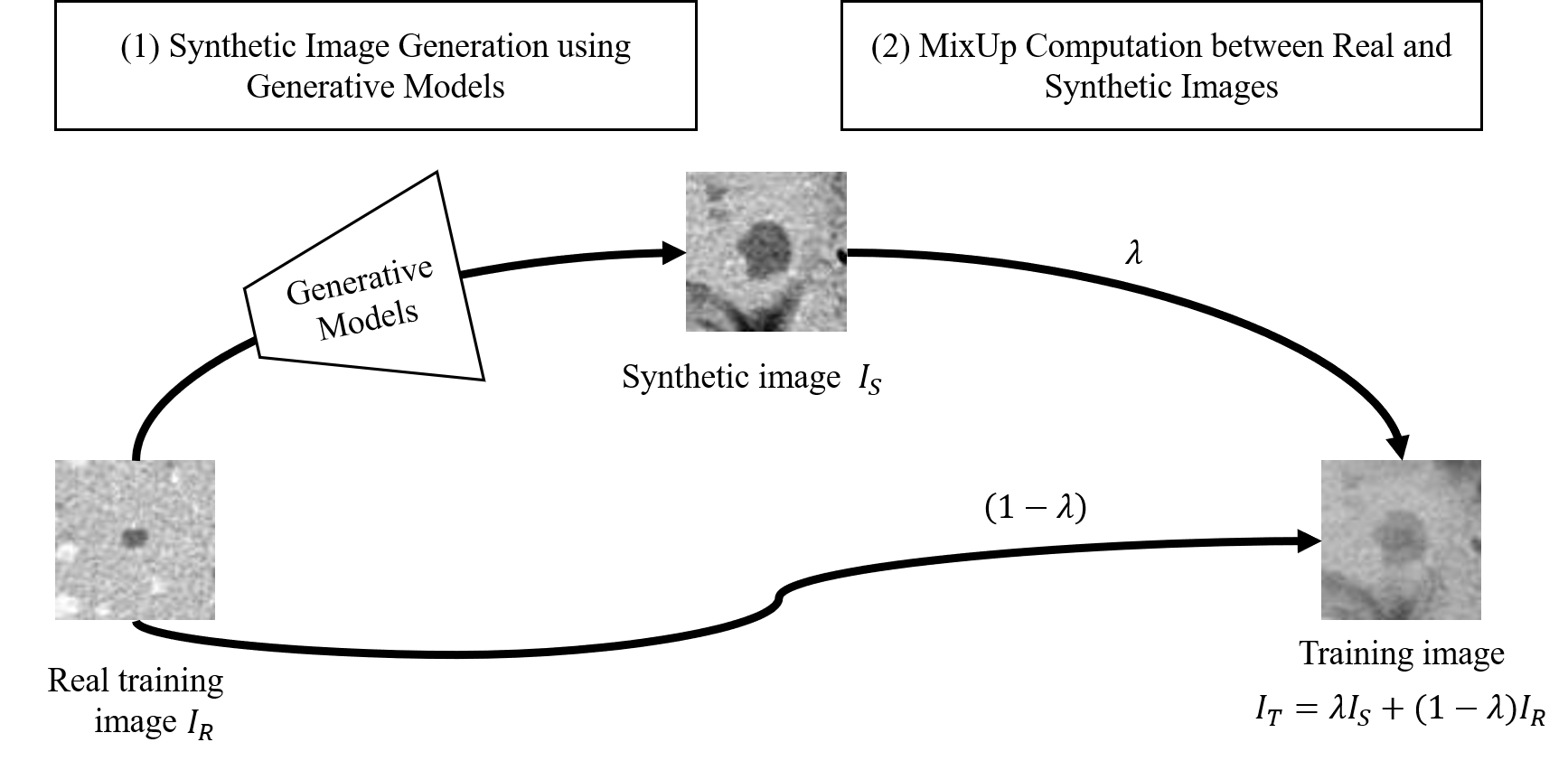} 
\caption{An overview of the proposed data augmentation method.}
\label{fig:method}
\end{figure*}

The proposed method consists of two main steps.
First, synthetic images of FLLs are generated from various generative models such as DCGAN, StyleGAN, Textual Inversion, and Diffusion Model.
Second, MixUp is applied between real image data and the mixed data of real and synthetic images.
Fig.~\ref{fig:method} summarizes an overview of the proposed method.

\subsection{Synthetic Image Generation with GANs and Diffusion Models}


In our proposed method, we use a generative model to generate synthetic images, specifically using DCGAN~\citep{Radford2015DCGAN}, StyleGAN~\citep{Karras2018StyleGAN}, Diffusion-based Textual Inversion~\citep{Gal2023TextualInversion}, and Diffusion Model~\citep{Dhariwal2021ADM}, which are state-of-the-art models capable of generating synthetic images. 

\noindent
\textbf{StyleGAN.}
The StyleGAN architecture consists of a mapping network and a synthesis network, with a novel style mixing technique for blending styles of two different images to produce a greater degree of variation in the generated images. 

From a data augmentation perspective, generative models offer several advantages for synthetic image creation. 
First, GANs can generate a wide range of diverse images that may not be present in the original dataset, which can be particularly useful in scenarios where the dataset is limited or biased towards certain types of images. 
Second, generating synthetic images using GANs is generally less expensive and time-consuming than manually collecting and annotating new images.
However, there are limitations to data augmentation through generative models. 
First, while GANs can generate realistic images, the quality of the generated images may not be as high as that of real images, which can limit the effectiveness of the synthetic images in training machine learning models. 
Second, GANs can be biased towards certain features or patterns that are present in the original dataset, and can suffer from mode collapse, where the generator network produces a limited set of output images that do not fully capture the diversity and complexity of the original image dataset. 
These limitations should be taken into account when using synthetic images for data augmentation in machine learning.

\noindent
\textbf{Textual Inversion.}
The Textual Inversion is designed to utilize text-to-image generation diffusion models without retraining them on new datasets. 
It introduces "textual tokens," which serve as placeholders for specific subjects or objects within the model's pre-existing vocabulary. 
During the optimization process, these tokens are fine-tuned to represent unique entities, allowing the model to generate images of these entities when provided with text prompt. 
This method leverages the robustness and generalizability of large-scale text-to-image models while providing a flexible and efficient means of generating customized images, tailored to represent specific subjects or styles not directly covered in the model's original training data.

In the proposed method, we employ Textual Inversion to make Latent Diffusion Model (LDM)~\citep{Rombach2022LDM} pre-trained on LAION-400B~\citep{Schuhmann2021LAION} to generate FLL images without additional training.
The Textual Inversion optimizes unique textual tokens serving as placeholders, which, through optimization, are adjusted to represent the specific characteristics of FLLs within the pre-trained framework of the diffusion model. 
Consequently, when these optimized tokens are included in a text prompt, the model generates synthetic images that mimic the desired features of FLLs, thereby enabling a customized data augmentation strategy without the computational overhead and data requirements typically associated with model finetuning.
The primary advantage of generating synthetic FFL images using Textual Inversion lies in its efficiency and flexibility. 
By leveraging the capabilities of a pre-trained diffusion model, it provides a practical solution for augmenting FFL datasets without the need for extensive computational resources and large datasets.
However, the quality and diversity of the generated images depend on the capabilities and biases of the pre-trained model. 
If the model has not encountered sufficient variations of FLL-like images during its initial training, the synthetic images may not fully capture the range of appearances that real FLLs can exhibit. 

\begin{figure*}[t!]
\centering
\begin{subfigure}{\textwidth}
\includegraphics[width=\linewidth]{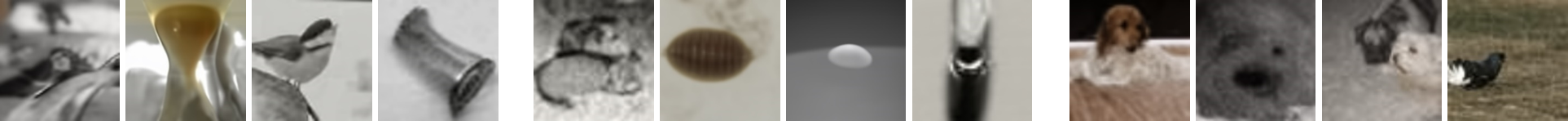} 
\caption{5k iterations}
\end{subfigure}
\begin{subfigure}{\textwidth}
\includegraphics[width=\linewidth]{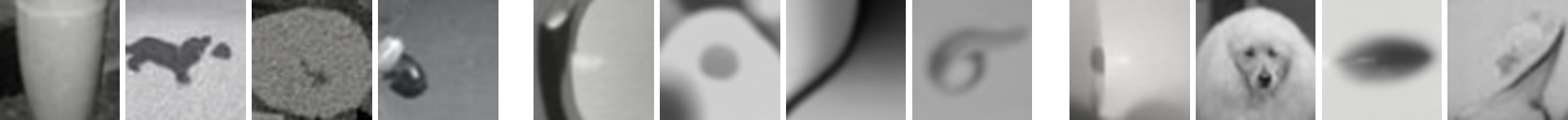}
\caption{10k iterations}
\end{subfigure}
\begin{subfigure}{\textwidth}
\includegraphics[width=\linewidth]{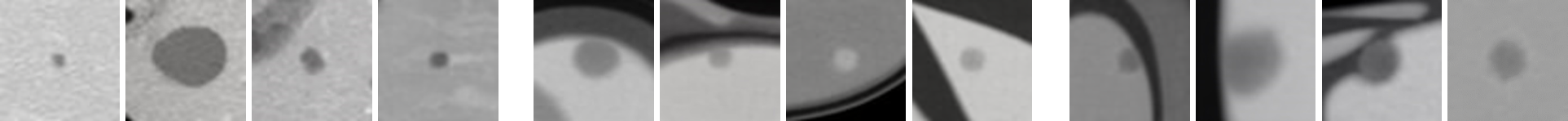}
\caption{15k iterations}
\end{subfigure}
\begin{subfigure}{\textwidth}
\includegraphics[width=\linewidth]{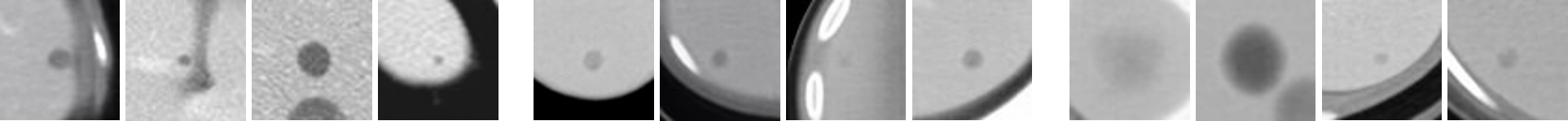}
\caption{20k iterations}
\end{subfigure}
\caption{The evolution of synthetic FLL images during the training process of the diffusion model. Initially, starting from an ImageNet-pretrained model, natural images are generated, which gradually converge to FLL patch images cropped from CT scans.}
\label{FIG:DM_EX}
\end{figure*}

\noindent
\textbf{Diffusion Model.}
Diffusion models are a class of generative models that generate data by iteratively denoising a variable that is initially pure noise. 
The model is trained through a process that involves gradually adding noise to training data and then learning to reverse this noising process. 
The primary components of a diffusion model include a forward process that adds noise to data in a controlled manner and a reverse process that learns to denoise the data step by step. 
The main advantage of diffusion models is their robustness in generating high-fidelity images that capture intricate details, which is crucial for medical image analysis. 

In the proposed method, we trained ImageNet-pretrained ADM diffusion model~\citep{Dhariwal2021ADM} on a training FLL dataset, and generated the synthetic images.
Fig.~\ref{FIG:DM_EX} shows the evolution of the synthetic FLL images generative by ADM during the training process.
Initially, the generated images were blurry and lacked distinct features, typical of early training phases in diffusion models. 
As training progressed, the images began to exhibit clearer structures and more defined features, indicating that the model was learning the underlying patterns of FLL images. 
By the final stages of training, the synthetic images were nearly indistinguishable from real FLL images.

\subsection{MixUp between Differently Sampled Data}


Generative models have succeeded in generating new patterns of images, but each comes with various limitations. 
As shown in Fig.~\ref{FIG:FLL_EX}, StyleGAN suffers from mode collapse, where similar patterns repetitively appear. 
Textual Inversion generates images that are dissimilar to actual FLL images, creating low quality visuals. 
Diffusion Models, while capable of creating diverse patterns similar to real images, tend to produce images with blurred details.
To address this issue, we propose applying the MixUp method to enable robust learning and improve the quality of the generative images.

In the proposed method, we form a mixed training data $(x_{T},y_{T})$ by mixing the synthetic data $(x_{S},y_{S})$ with the original real training data $(x_{R},y_{R})$ as follows:

\begin{equation}
    x_{T} = \lambda x_{S} + (1-\lambda) x_{R},~y_{T} = \lambda y_{S} + (1-\lambda) y_{R},
    \label{eq:5}
\end{equation}

\noindent
where $\lambda$ is the MixUp coefficient determined as $\lambda \sim \text{Beta}(\alpha,1)$.

The proposed method of calculating MixUp between synthetic and real images can yield several benefits. 
First, if the quality of the synthetic images is low or there are discrepancies compared to the real images, MixUp can generate data that mitigates this sense of heterogeneity. 
Second, when synthetic images are biased towards a specific pattern, MixUp can generate data with various patterns and appearances by mixing them with actual images featuring relatively diverse patterns and appearances.

%% file: sections/4_experiments.tex
\section{Experiments}
\label{SEC:EXPERIMENTS}

\subsection{Experimental Settings}
\label{SEC:EXP_SETTING}

\noindent
\textbf{Datasets.}
We conducted an experiment on the task of classifying diseases of FLLs on abdominal CT images~\citep{Bae2021,Lee2021}. 
The dataset consists of CT scans collected from 502 colorectal cancer patients, including 1,290 focal liver lesion patches with 676 cysts, 130 hemangiomas, and 484 metastases. 
All images have a resolution of $512\times 512$, a pixel size in the range of $0.5 \times 0.5$ mm\textsuperscript{2} to $0.8 \times 0.8$ mm\textsuperscript{2}, and slice thickness between 3 to 5 mm. 
All FLLs are manually segmented on the axial plane with the largest cross section and the dataset is split into a training set of 681 lesions (433C+70H+178M), a validation set of 302 lesions (115C+30H+157M), and a test set of 307 lesions (128C+30H+149M) based on the acquisition date.

\noindent
\textbf{Comparison and Evaluation.}
To verify the effectiveness of the proposed method, we compared the results of the proposed method with the results of the following data augmentation settings; (1) Real data without augmentation (Baseline), (2) two mixture data augmentation with real data (MixUp, AugMix), (3) blended data of real and synthetic data (DCGAN, StyleGAN, Textual Inversion, Diffusion Model), and (4) the proposed method with various generative models. 
The experimental results were evaluated both quantitatively and qualitatively.
First, we compared examples of real, synthetic, and GenMix images to assess the quality of synthetic images.
Second, the performance of FLL classification were evaluated and compared by measuring accuracy, F1 score, and sensitivity and specificity for each class.
It is noteworthy that the F1 score is regarded as a more suitable metric for class imbalanced data than the accuracy by emphasizing the minor class performance.
Lastly, we qualitatively analyzed how the generative models and the GenMix contribute to learning by examining the tSNE distribution of features extracted from the classifiers.

\noindent
\textbf{Implementation Details.}
For training DCGAN and StyleGAN, we set the hyperparameters to 70,000 iterations, a batch size of 8, and a learning rate of 0.001. 
The number of generated images was set to be equal to the number of training images for each class. 
In synthetic image generation stage, we employed the LDMs~\citep{Rombach2022LDM} pre-trained on the LAION-400M dataset~\citep{Schuhmann2021LAION} for applying Textual Inversion.
In MixUp stage of the proposed method, we set the MixUp parameter to $\alpha=0.3$. 
We used the ImageNet-pretrained VGG-16~\citep{Simonyan2014VGG} network for the FLL classification. 
For training VGG-16, we set the hyperparameters to 100 epochs, a batch size of 8, a learning rate of 0.0002, and an early stopping condition of 20.

\subsection{Results}
\label{SEC:RESULT}

\begin{figure*}[t!]
\centering
\begin{subfigure}{\textwidth}
\includegraphics[width=\linewidth]{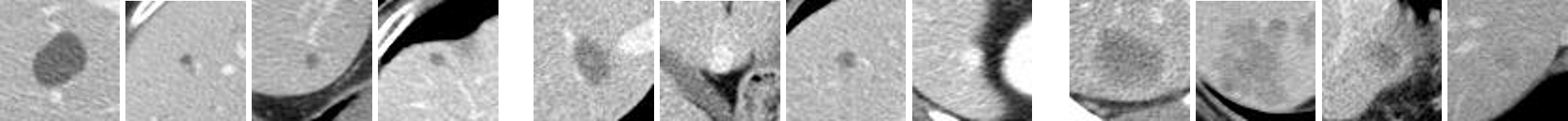} 
\caption{Real images}
\end{subfigure}
\begin{subfigure}{\textwidth}
\includegraphics[width=\linewidth]{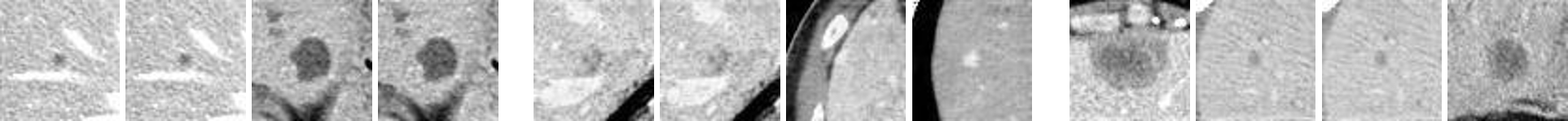}
\caption{Synthetic images (StyleGAN)}
\end{subfigure}
\begin{subfigure}{\textwidth}
\includegraphics[width=\linewidth]{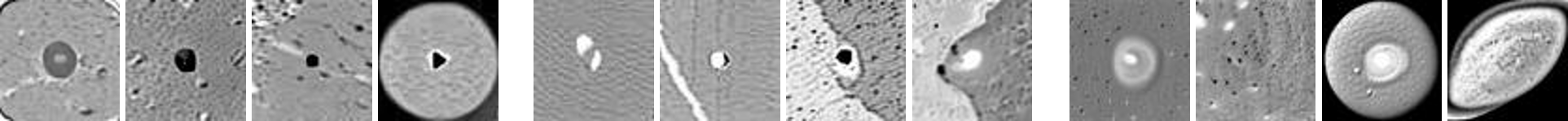}
\caption{Synthetic images (Textual Inversion)}
\end{subfigure}
\begin{subfigure}{\textwidth}
\includegraphics[width=\linewidth]{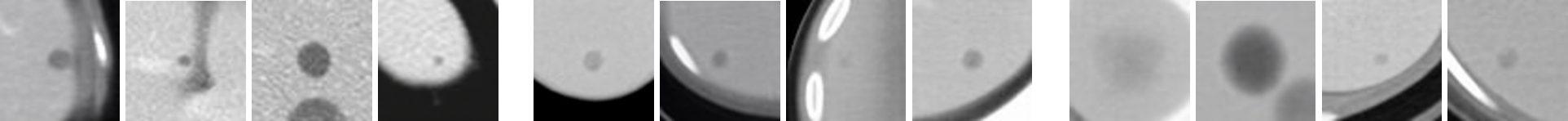}
\caption{Synthetic images (Diffusion Model)}
\end{subfigure}
\begin{subfigure}{\textwidth}
\includegraphics[width=\linewidth]{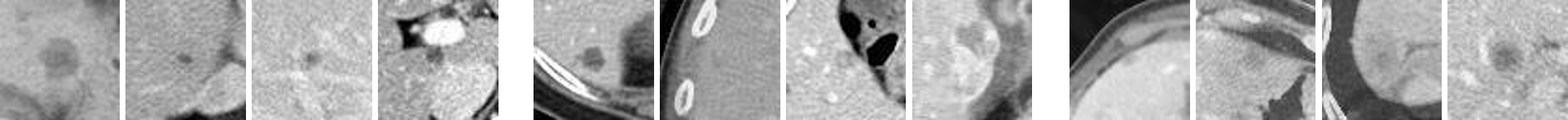}
\caption{GenMix training images (StyleGAN)}
\end{subfigure}
\begin{subfigure}{\textwidth}
\includegraphics[width=\linewidth]{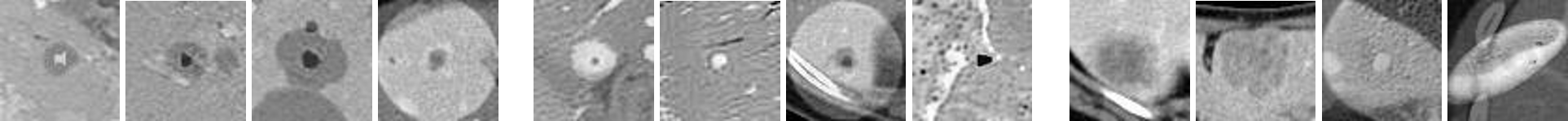}
\caption{GenMix training images (Textual Inversion)}
\end{subfigure}
\begin{subfigure}{\textwidth}
\includegraphics[width=\linewidth]{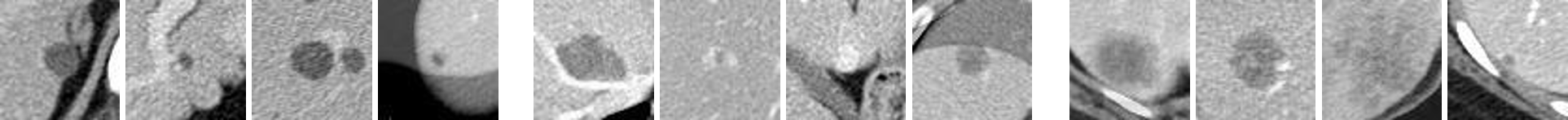}
\caption{GenMix training images (Diffusion Model)}
\end{subfigure}
\caption{Examples of FLL images for cysts (left), hemangiomas (middle), and metastases (right) in real, synthetic, and GenMix training data.}
\label{FIG:FLL_EX}
\end{figure*}

\noindent
\textbf{Image Assessment.}
Fig.~\ref{FIG:FLL_EX} shows examples of real images, synthetic images generated by SytleGAN, Textual Inversion, and diffusion model, and their GenMix images for each class.
In real images, cysts typically exhibit dark and homogeneous intensity values, hemangiomas are generally characterized by the presence of intensity-enhanced blood vessels surrounding them, and metastases have indistinct boundaries with inhomogeneous internal intensity. 
While the appearance characteristics between diseases are relatively distinct in larger FLLs, they become less distinguishable in small-sized lesions.

The synthetic images generated by StyleGAN not only accurately reflect these disease characteristics of real images but also produce high-quality images remarkably similar to real ones. 
However, due to the limited amount of data and patterns, a limitation of the GAN technique, known as mode collapse, was observed, resulting in the repeated generation of images with limited patterns. 
On the other hand, synthetic images generated through Textual Inversion showed a wider variety of patterns compared to those generated by StyleGAN. 
Interestingly, despite not being trained on FLL images, synthetic images reflecting the appearance characteristics of cysts with dark and homogeneous intensity and hemangiomas with enhanced blood vessels were generated. 
Nevertheless, due to the absence of fine-tuning on the FLL dataset, there are limitations in the quality of generated images, which exhibit a certain degree of discrepancy from real images upon visual inspection.
Lastly, images generated by the Diffusion Model fine-tuned on the FLL dataset exhibit a much closer resemblance to real FLL images compared to those generated by Textual Inversion, and show far more diverse patterns than those produced by StyleGAN. 
This indicates that the Diffusion Model is more robust against mode collapse and is capable of generating higher-quality images than GAN-based techniques. 
However, due to the limited training data, some generated images display slightly abnormal patterns or appear blurred, indicating a loss of image details.

The GenMix images show that the quality of synthetic images generated from generative models is improved regardless of the type of generative models.
For StyleGAN, which generates limited pattern images due to mode collapse, the GenMix-StyleGAN images show increased pattern diversity through mixup with real images. 
In the case of Textual Inversion, which produces diverse but low-quality images that are dissimilar to actual images, the GenMix-Textual Inversion images show improved quality, making them much closer to real images. 
Although Diffusion Models initially produced images with somewhat blurred appearances, the GenMix-Diffusion Model images exhibit significantly enhanced image details. 
This observation confirms that the proposed GenMix enhances the quality of synthetic images and addresses their limitations through mixup with real images.


\begin{table*}[t!]
\caption{Performance comparison for various data augmentation methods. The numbers in parentheses of methods refer to the ratio of synthetic images to training data. (Acc.: Accuracy, F1: F1 score)}
\centering
\begin{tabular}{L{4.2cm} | C{0.8cm} C{0.8cm} | C{0.8cm} C{0.8cm} | C{0.8cm} C{0.8cm} | C{0.8cm} C{0.8cm} }
\hline
 & & & \multicolumn{2}{c}{Cyst} & \multicolumn{2}{c}{Hemangioma} & \multicolumn{2}{c}{Metastasis} \\
Methods & Acc. & F1 & Sens. & Spec. & Sens. & Spec. & Sens. & Spec. \\
\hline \hline
Baseline (Real) & 72.3 & 60.7 & 85.6 & \textbf{82.0} & 29.3 & 89.5 & 69.4 & 84.9 \\
MixUp (Real) & 74.1 & 61.7 & 89.4 & 76.7 & 23.3 & \textbf{96.0} & 71.3 & 83.2 \\
AugMix (Real) & 70.5 & 60.9 & 85.6 & 78.6 & 37.3 & 89.4 & 64.2 & \textbf{85.6} \\
\hline
DCGAN (50\%) & 72.0 & 60.0 & 89.1 & 59.8 & 26.7 & 76.9 & 66.4 & 77.2 \\
StyleGAN (40\%) & 75.9 & 63.9 & 87.5 & 67.6 & 30.0 & 80.9 & 75.2 & 76.6 \\
Textual Inversion (50\%) & 72.0 & 60.5 & 60.9 & 79.9 & 23.3 & 77.3 & \textbf{91.3} & 53.8 \\
Diffusion Model (50\%) & 75.6 & 65.6 & 89.8 & 65.4 & \textbf{40.0} & 79.4 & 70.5 & 80.4 \\
\hline
\textbf{GenMix (DCGAN)} & 76.9 & 63.4 & \textbf{93.8} & 64.8 & 23.3 & 82.7 & 73.2 & 80.4 \\
\textbf{GenMix (StyleGAN)} & 80.8 & 69.6 & \textbf{93.8} & 71.5 & 30.0 & 86.3 & 79.9 & 81.6 \\
\textbf{GenMix (Textual Inversion)} & \textbf{81.4} & \textbf{70.6} & 91.4 & 74.3 & 33.3 & 86.6 & 82.6 & 80.4 \\
\textbf{GenMix (Diffusion Model)} & 79.8 & 69.3 & 86.7 & 74.9 & 36.7 & 84.5 & 82.6 & 77.2 \\
\hline
\end{tabular}
\label{TAB:PERF_EVAL}
\end{table*}

\noindent
\textbf{Performance Evaluation.}
Table~\ref{TAB:PERF_EVAL} summarizes the quantitative evaluation of FLL classification for the proposed method and comparative methods.
The baseline model generally shows high sensitivity in the major classes, cyst and metastasis, while the minor class, hemangioma, has low sensitivity due to class imbalance.
When MixUp was applied to real images, an improvement was observed in the sensitivity for cysts and metastases, resulting in enhanced accuracy and F1 scores.
The results where the real-image-based baseline and mixture data augmentation each achieved the highest specificity for all three classes are noteworthy.

When applying GAN-based data augmentation, DCGAN failed to improve the original classification performance. 
In contrast, StyleGAN, despite generating a limited pattern of images, improved the sensitivity for metastases through the generation of high-quality synthetic images, enhancing both accuracy and F1 scores by approximately 3\%p. 
Due to the degradation in image quality, Textual Inversion showed minimal overall improvement in classification performance. 
When examining class-specific performance, there was a significant decrease in the sensitivity for cysts, which was counterbalanced by an increase in the sensitivity for metastasis. 
On the other hand, the Diffusion Model generated relatively high-quality images with diverse patterns, resulting in the highest F1 score among the generative models and significantly improved sensitivity for hemangiomas. 
A common characteristic observed with generative model-based data augmentation is the decreased specificity for each class compared to the real image-based baseline and mixture data augmentation.

The results of GenMix exhibit three notable characteristics. 
First, GenMix improved the accuracy and F1-score of all generative models. 
This improvement can be attributed to GenMix enhancing the quality of synthetic images through mixup with real images and strengthening class boundary information through mixup among real images. 
Second, GenMix improved the class-specific specificity of the generative models. 
Finally, among the generative models, Textual Inversion achieved the highest performance with GenMix. 
This is particularly interesting because, despite its limitation of producing low-quality images and thus barely improving classification performance on its own, the proposed GenMix compensated for this quality issue, allowing the diversity of Textual Inversion to shine. 
Moreover, since Textual Inversion was not fine-tuned on the FLL dataset, this result is especially noteworthy.

\begin{figure*}[t!]
\centering
\begin{subfigure}{0.44\textwidth}
\includegraphics[width=\linewidth]{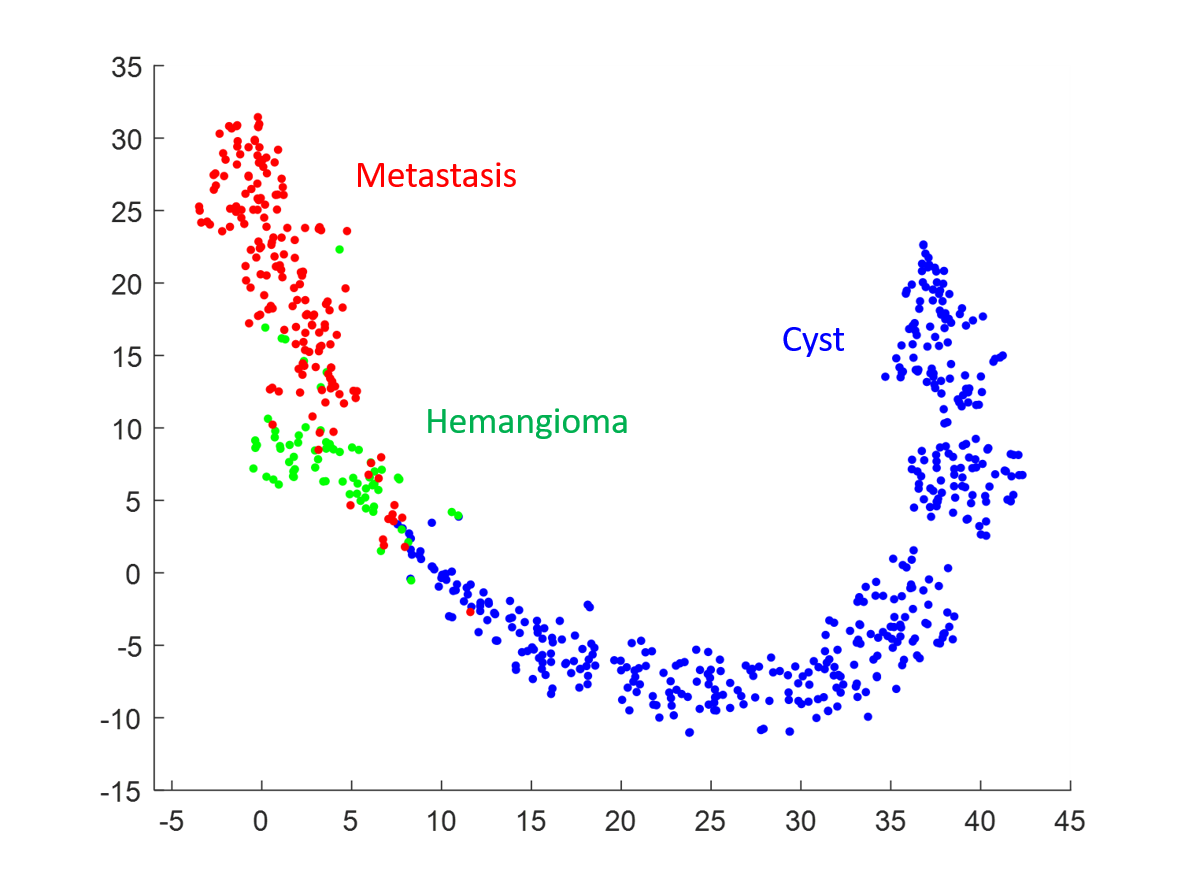}
\caption{Baseline (Real Only)}
\end{subfigure}
\\
\begin{subfigure}{0.425\textwidth}
\includegraphics[width=\linewidth]{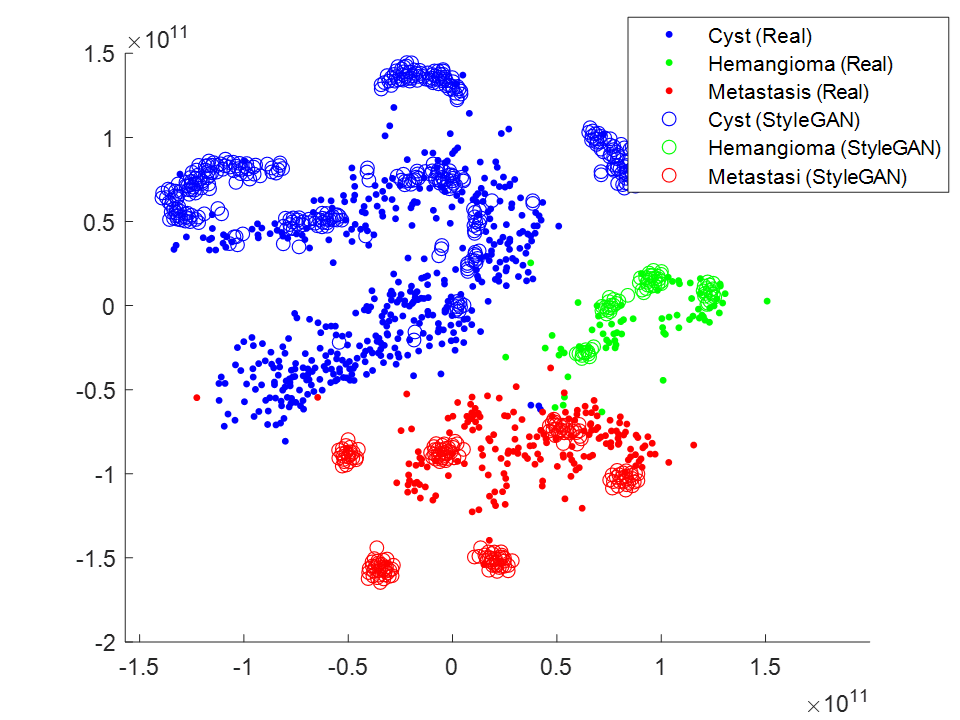}
\caption{StyleGAN}
\end{subfigure}
\begin{subfigure}{0.425\textwidth}
\includegraphics[width=\linewidth]{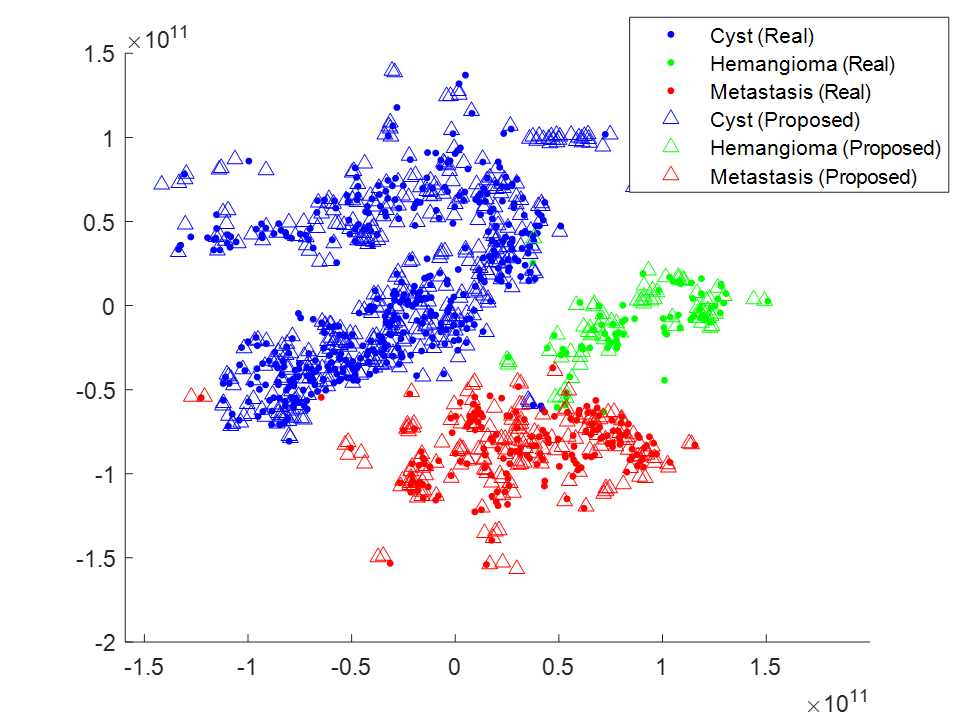}
\caption{GenMix (StyleGAN)}
\end{subfigure}
\begin{subfigure}{0.425\textwidth}
\includegraphics[width=\linewidth]{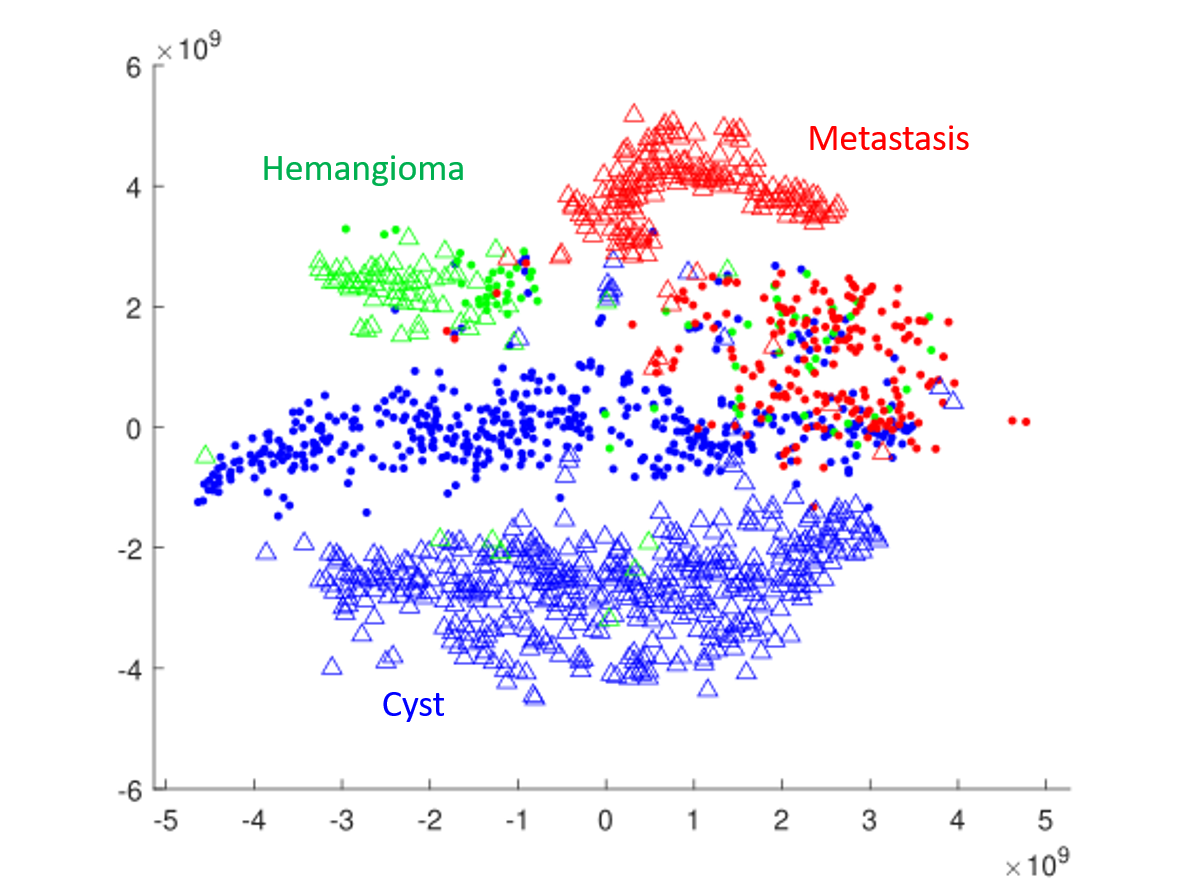}
\caption{Textual Inversion}
\end{subfigure}
\begin{subfigure}{0.425\textwidth}
\includegraphics[width=\linewidth]{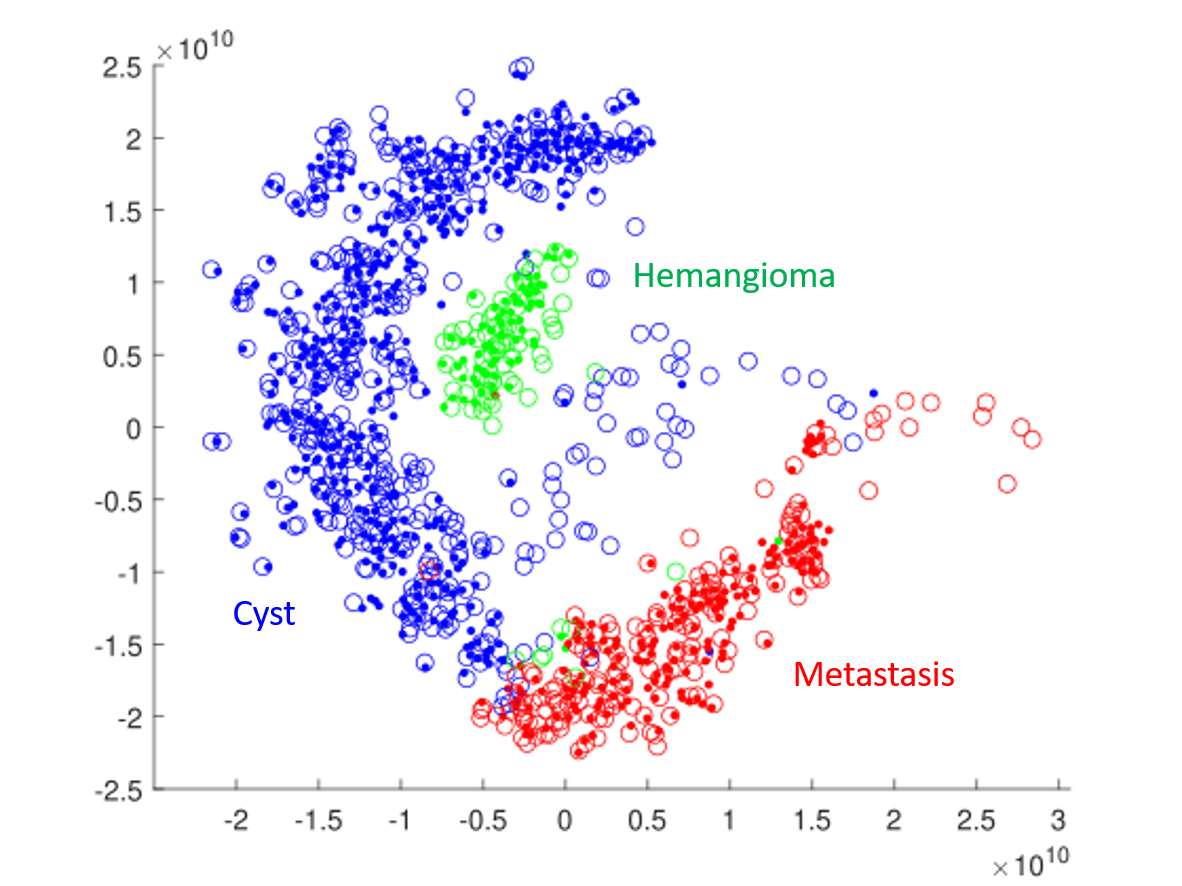}
\caption{GenMix (Textual Inversion)}
\end{subfigure}
\begin{subfigure}{0.425\textwidth}
\includegraphics[width=\linewidth]{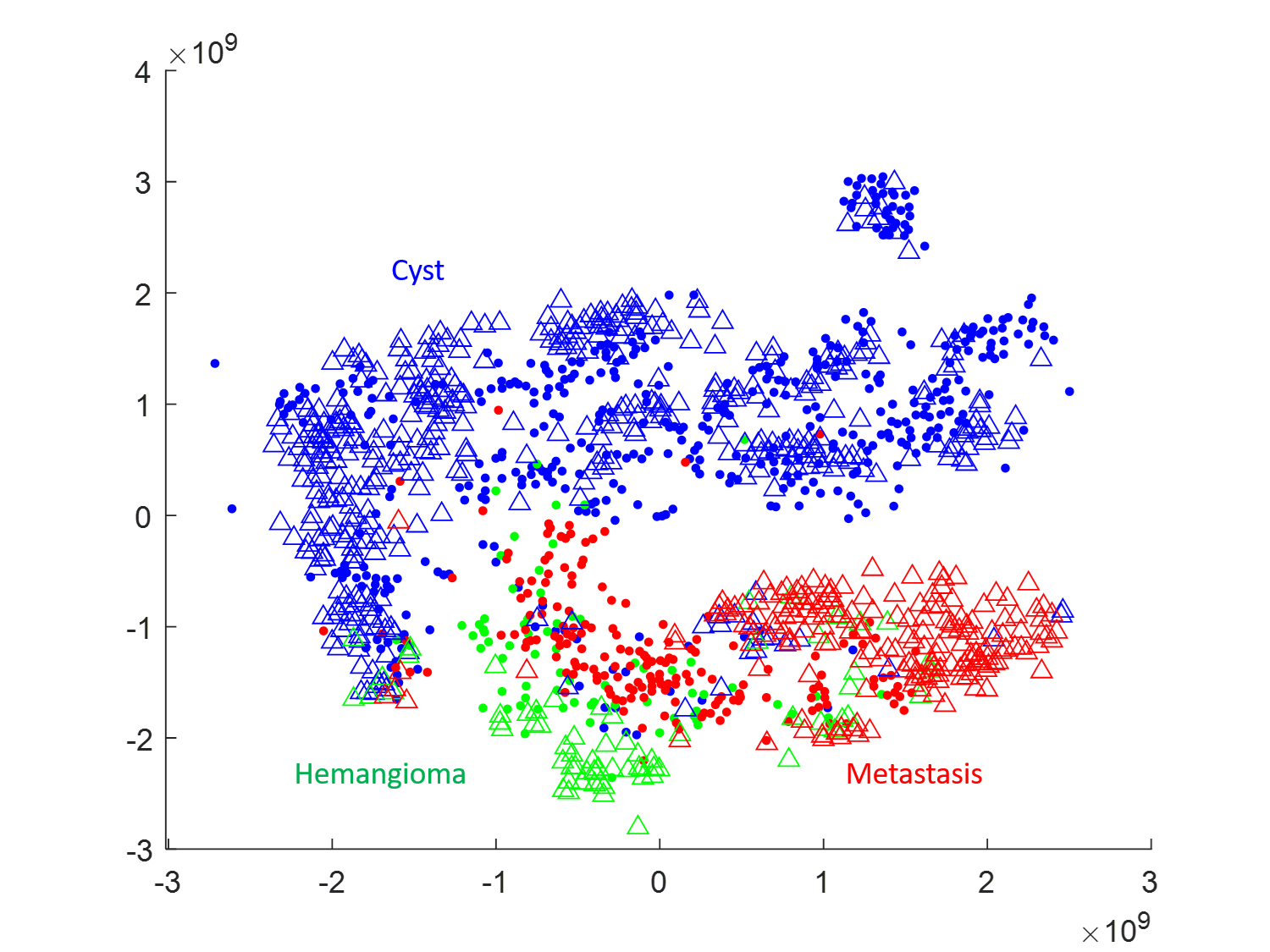}
\caption{Diffusion Model}
\end{subfigure}
\begin{subfigure}{0.425\textwidth}
\includegraphics[width=\linewidth]{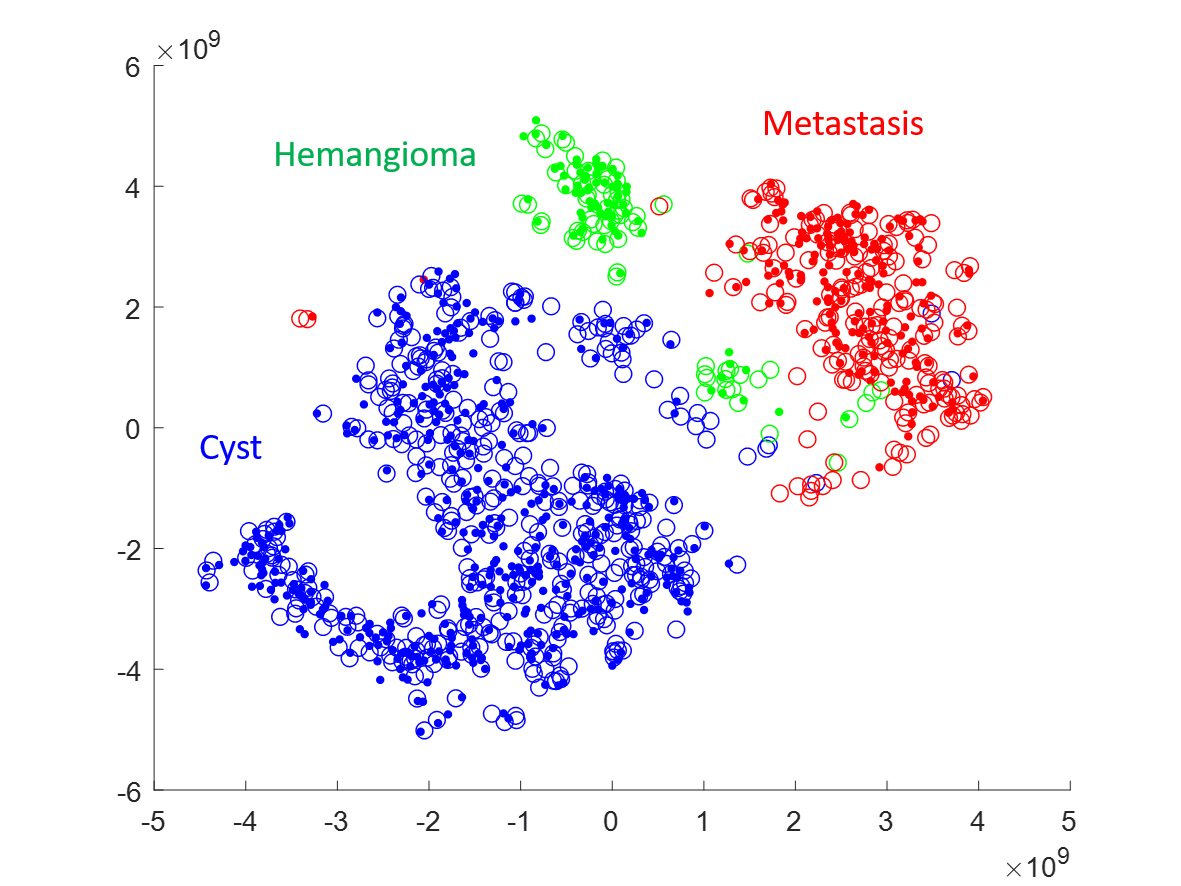}
\caption{GenMix (Diffusion Model)}
\end{subfigure}
\caption{t-SNE visualization of the distribution of training data features extracted from VGG-16 trained on various data augmentation methods. (\textcolor{blue}{Blue}: Cyst, \textcolor{green}{Green}: Hemangioma, \textcolor{red}{Red}: Metastasis, Dots: Real data, Triangles: Synthetic data, Circles: GenMix data)}
\label{FIG:TSNE}
\end{figure*}

\noindent
\textbf{tSNE Feature Analysis.}
Fig.~\ref{FIG:TSNE} presents the t-SNE visualization of feature distributions extracted from the VGG-16 networks trained on various training data.
The tSNE distribution of real data reveals that the two major classes, cysts and metastases, are closely situated, with the minor class, hemangioma, distributed along their boundary area. 
To accurately classify these, data augmentation must be performed to (1) strengthen the in-class distribution of hemangioma and (2) reinforce the boundary areas between the three classes. 

The tSNE distribution of StyleGAN in Fig.~\ref{FIG:TSNE} (b) indicates that the synthetic data, marked by rectangles, is distributed (1) across the outer regions of each class's distribution and (2) clustered in specific sections. 
The first feature signifies that StyleGAN generates images with pronounced class-specific features, particularly in the outer regions rather than the boundary areas where class features are similar. 
The second feature implies that StyleGAN exhibits the mode collapse phenomenon, repetitively generating images of specific patterns. 
Although this strengthens each class's distribution, the effect of data augmentation on the boundary areas between classes is minimal, leading to a marginal improvement in classification.
In contrast, the t-SNE distribution of GenMix in Fig.~\ref{FIG:TSNE} (c) reveals that the GenMix data is evenly distributed across the entire class range of real data, rather than being concentrated in specific areas. 
This indicates that through mixup between StyleGAN data and real images, the distribution has expanded across the entire class spectrum, enhancing the diversity of the synthetic data. 
This increased diversity is also reflected in the feature distribution.

The tSNE distribution of Textual Inversion in Fig.~\ref{FIG:TSNE} (d) shows that the synthetic data is closely aligned with the real data's distribution for all classes but remains distinctly separated. 
This suggests that while Textual Inversion generates data capturing some characteristics of each class from the real data, the absence of fine-tuning for real data causes a mean shift in the distributions, keeping them apart.
Conversely, the tSNE distribution of GenMix in Fig.~\ref{FIG:TSNE} (e) reveals that GenMix data, marked by circles, almost overlaps with the real data distribution, significantly covering the boundary areas between classes. 
This can be interpreted as the MixUp between real and synthetic data providing an effective fine-tuning effect that merges the two separate distributions. 

The t-SNE distribution of the Diffusion Model in Fig.~\ref{FIG:TSNE} (f) reveals that, unlike StyleGAN or Textual Inversion, the synthetic data distribution nearly overlaps with the real data distribution. 
This indicates that the Diffusion Model is robust against model collapse and generates high-quality data similar to real images. 
However, the generated data is rarely distributed in the boundary regions between classes. 
This limitation is inherent to generative models, as they tend to learn salient features of the target class, resulting in minimal generation of data in ambiguous boundary regions. 
In contrast, the t-SNE distribution of GenMix in Fig.~\ref{FIG:TSNE} (g), which involves mixup between synthetic and real images, shows that several GenMix data points are also distributed in the class boundary regions. 
This demonstrates that GenMix not only improves the quality of synthetic data but also enhances boundary region information, thereby improving the efficiency of medical image analysis.

%% file: sections/6_conclusion.tex
\section{Conclusion}
\label{SEC:CONCLUSION}

In this paper, we proposed GenMix, a data augmentation technique that combines generative and mixture approaches to leverage their strengths.
Generative models create new data patterns but face issues like mode collapse and training difficulties, especially with limited medical imaging data. 
Mixture models enhance class boundaries but can be biased towards major classes. 
GenMix addresses these limitations by integrating both approaches in two stages: (1) training a generative model to produce synthetic images, and (2) performing mixup between synthetic and real data.
Experiments validated GenMix's effectiveness in classifying focal liver lesions (FLL) in CT images. 
GenMix improved the accuracy and F1-score of various generative models, including DCGAN, StyleGAN, Textual Inversion, and Diffusion Models. 
Mixup with real images enhanced the quality and diversity of synthetic images and strengthened class boundary regions. 
Notably, Textual Inversion achieved the highest performance with GenMix, highlighting its potential to compensate for generative models' shortcomings without additional fine-tuning on the FLL dataset.
Future research should explore GenMix's application to other medical imaging tasks and datasets to validate its generalizability. 
GenMix promises to enhance machine learning models' performance in medical image analysis, leading to more accurate and reliable diagnostic tools.